\documentclass[letterpaper, 10 pt, conference]{ieeeconf}  % Comment this line out if you need a4paper

\IEEEoverridecommandlockouts                              % This command is only needed if 
\let\labelindent\relax
\usepackage{amsmath} % assumes amsmath package installed
\usepackage{amssymb}  % assumes amsmath package installed

\usepackage{xspace}
\usepackage{color}
\usepackage{xcolor}
\usepackage[colorlinks]{hyperref}
\usepackage{enumitem} % enumeration
\usepackage{bm}
\usepackage{graphicx}
\usepackage{booktabs}
\usepackage{multirow}
\newcommand{\eg}{\textit{e.g.}\xspace}

\title{\LARGE \bf
Vision-based Vineyard Navigation Solution with Automatic Annotation
}

\author{Ertai Liu, Josephine Monica, Kaitlin Gold, Lance Cadle-Davidson, David Combs, Yu Jiang% <-this % stops a space
\thanks{Ertai Liu and Yu Jiang are with the Horticulture Section, SIPS, Cornell AgriTech, Geneva, NY 14456, USA and the Department of Biological and Environmental Engineering, Cornell University, Ithaca, NY 14850, USA. Josephine Monica is with the Sibley School of Mechanical and Aerospace Engineering, Cornell University, Ithaca, NY 14850, USA. Kaitlin Gold and David Combs are with the PPPMB Section, SIPS,
Cornell AgriTech, Geneva, NY 14456, USA. Lance Cadle-Davidson is
with the USDA ARS Grape Genetics Research Unit, Geneva, NY, 14456,
USA.
        {\tt\small \{el766, jm2684, kg557, dbc10, yujiang\}@cornell.edu and lance.cadledavidson@usda.gov}
}%
}

\begin{document}

\maketitle
\thispagestyle{empty}
\pagestyle{empty}

%%%%%%%%%%%%%%%%%%%%%%%%%%%%%%%%%%%%%%%%%%%%%%%%%%%%%%%%%%%%%%%%%%%%%%%%%%%%%%%%

%%%%%%%%%%%%%%%%%%%%%%%%%%%%%%%%%%%%%%%%%%%%%%%%%%%%%%%%%%%%%%%%%%%%%%%%%%%%%%%%
\begin{abstract}
Autonomous navigation is the key to achieving the full automation of agricultural research and production management (e.g., disease management and yield prediction) using agricultural robots. In this paper, we introduced a vision-based autonomous navigation framework for agriculture robots in trellised cropping systems such as vineyards.
To achieve this, we proposed a novel learning-based method to estimate the \emph{path traversibility heatmap} directly from an RGB-D image and subsequently convert the heatmap to a preferred traversal path.
An \emph{automatic annotation pipeline} was developed to form a training dataset by projecting RTK GPS paths collected during the first setup in a vineyard in corresponding RGB-D images as ground-truth path annotations, allowing a fast model training and fine-tuning without costly human annotation.
The trained path detection model was used to develop a full navigation framework consisting of row tracking and row switching modules, enabling a robot to traverse within a crop row and transit between crop rows to cover an entire vineyard autonomously.
Extensive field trials were conducted in three different vineyards to demonstrate that the developed path detection model and navigation framework provided a cost-effective, accurate, and robust autonomous navigation solution in the vineyard and could be generalized to unseen vineyards with stable performance.
\end{abstract}
\section{Introduction}
Viticulture has been known as a labor-intensive production throughout its 8,000-year history. The development and application of agriculture machinery and pesticides have changed the picture of fieldwork by providing effective assistance in the field management tasks, and revealed the demand of further reducing the agricultural labor requirement by moving toward the next generation full automation of fieldwork. 

Many intelligent actuation systems have been developed for automating the planting, pruning, and harvesting \cite{ly2015fully,botterill2017robot,luo2018vision}. Recently the exploration of novel robotic perception system for disease management and yield prediction tasks have also gained attention because of the development of modern computer vision technology \cite{liu2022near,lopes2016vineyard,victorino2019grapevine}. In addition to these task-oriented automation, robotic systems need to accurately and autonomously navigate in the field with adaption to ever-changing field conditions. Currently, navigation systems based on real time kinematics (RTK)-enabled global navigation satellite system (GNSS) and inertial measurement unit (IMU) have been widely used in both research projects and commercial applications. However, RTK-GPS based systems heavily rely on fast and stable communication (e.g., radio) with a base station. When communication interference occurs (e.g., poor visibility to satellites due to tall crops), the RTK-GPS signals will be interrupted, influencing or even ceasing robot operations. Moreover, if a mobile base station is used and moved unexpectedly, the rover localization will be dramatically changed, leading to potential safety risks such as a collision in the field.

Vision-based navigation solutions have been extensively studied in recent days because of 1) increased imaging resolution and modalities (e.g., depth cameras) with affordable cost for resolving needed information in navigation and 2) breakthroughs in deep learning (DL) that offer new tools for image analysis and interpretation for navigation. Many recent vision-based navigation systems require abundant human annotations of certain objects in the field (e.g., crops and crop rows) to train a DL model for finding a traversal path. While these systems have achieved certain success, challenges remain in i) costly data annotation for model training and ii) substantial variations among crops (e.g., corns vs grapes) and cropping systems (e.g., row cropping vs trellised cropping systems) for generalizing a model/system developed from one crop to another. These become major barriers for adopting the vision-based navigation solutions in agriculture.

This work aimed to provide a learning-based visual navigation system that can be easily and reliably deployed in the field. We proposed to directly output the traversal path in image frames, and automatically generate the training annotation by projecting the RTK-GPS paths onto the images. The RTK GPS would be required only during the setup phase for training data collection and could be removed during the deployment phase for regular operations, reducing the cost of training and applying a DL for various crops and cropping systems. We integrated the pretrained model into a full navigation framework consisting of row tracking and switching modules for autonomous navigation in the vineyard.

\section{Related Work}
Previous studies on autonomous navigation for field robots~\cite{thuilot2002automatic, griepentrog2006autonomous, kayacan2018high, reid2000agricultural, cordesses2000combine} have primarily relied on GPS with RTK correction combined with IMU. This is due to the accuracy of GPS with RTK correction which can reach centimeter level accuracy. However, this approach is limited by the need of having reliable GPS signals, which may be difficult to obtain in certain circumstances (\eg on the area with dense vegetation and tall crops). In such failure cases, dead-reckoning using IMU alone may not be reliable, especially on the fields with uneven surfaces and slopes. Obtaining an RTK signal correction is also necessary for GPS-based navigation to reach high accuracy, which can be expensive and impractical in many situations. Additionally, GPS-based navigation also requires long preparation and surveying the field to obtain a pre-defined map of the field structures and determine the path ahead of time. Furthermore, this map may not always be reliable due to changes in terrain conditions, weather, and season. 
These limitations make it challenging to apply GPS-based navigation to new fields and environments, hindering its widespread use. 

Therefore, researchers have investigated other sensors to supplement or replace GPS. 
For instance, LiDAR has been utilized to provide point cloud spatial information that enables the leveraging of the structure of rows or plants for in-field navigation~\cite{higuti2019under}. 
Some LiDAR-based methods perform line fitting (\eg RANSAC,  Hough transform, and PEARL~\cite{isack2012energy}) to identify the plant row structures from the point cloud~\cite{barawid2007development,malavazi2018lidar,nehme2021lidar}. 
Others directly use the point cloud as measurements to filter-based localization (\eg particle filter) and navigate at the center of the crops~\cite{hiremath2014laser}.
%However, point cloud-based methods can be sensitive to  point cloud density and may not generalize well to different plants and fields, as the density and structure of crops can vary and thus affect the distribution of the point cloud.
Although LiDAR provides accurate spatial information, the state of the art LiDAR with high density is expensive and cost prohibitive for consumer products.
Moreover, point cloud-based methods may struggle to adapt to different plants and fields, as the density and structure of crops can vary and thus affect the distribution of the point cloud.
These methods are also sensitive to the stage (\eg early or late) of the crops and plant size. 
For example, crops in their early stages may have smaller and less prominent structures that are harder to detect with LiDAR. Indeed, these methods have only been shown to work well for relatively tall vegetation with prominent structures, such as vineyards and maize, and may not work well for row crops that lack prominent structures, such as carrots.
More importantly, LiDAR does not provide semantic information that can be useful for improving the robustness of the system.

%%%%%%%%%%%%%%%%%%%%%%%%%%%%%%
%\cite{barawid2007development} uses laser scanner to navigate orchard row crops by using Hough transform to detect orchard rows as straight line
%\cite{malavazi2018lidar} perform line extractions from 2D LiDAR point clouds using a PEARL~\cite{isack2012energy} geometric fitting based method and apply additional filter and post processing steps for crop detection and extract line path to follow.
%\cite{nehme2021lidar} applies Hough transform to point cloud input for line fitting in vineyard and identify the structure of the field.
%\cite{hiremath2014laser} uses particle filter and LiDAR measurement to localize the robot and navigate at the center of the crops. 
%%%%%%%%%%%%%%%%%%%%%

%Thus, many works leverage camera image information to perform vision-based navigation.
Thus, vision-based navigation, using image input from the camera for path guidance, has become increasingly popular in recent years.
Some of these works involve using image input from the camera to detect plant rows and lines for path guidance.
For instance, \cite{ahmadi2022towards, ahmadi2020visual} use classical image signal processing techniques (such as thresholding color values) to extract vegetation masks for row crop plants, which are then subjected to pixel clustering and post-processing steps to obtain a line for navigation guidance. 
While classical signal processing techniques may effectively extract vegetation masks for crop plants (short plants forming clear and simple lines), these techniques may not work well in more complicated plant and field scenarios.
Thus, deep learning-based techniques have also been used to address this challenge.
\cite{aghi2021deep} trains a canopy segmentation model MobileNetV3, and then performs pixel clustering on the segmentation results to obtain a cluster of empty space, which is identified as the road path.
However, this method relies on having only one empty space cluster, thus may not work well in several cases, \eg, early stages of growth or sparse vegetation. 
Moreover, the feedback on control commands is done in pixel space, which may not accurately represent the robot's movements. Working solely in pixel space limits the ability to sense distance and represent changes in elevation. This may result to errors in planning control commands, especially in complex environments such as varying terrain features that encounter slopes or elevation changes. Our proposed approach avoids relying explicitly on the crop patterns and adapts variant field environment with the generalizability of the machine learning model during the path detection. The detected path is transferred to the Bird Eye View (BEV) before applying the controller, and thus reduces the limitation of working soly on the image plane.
Alternatively, \cite{sivakumar2021learned} uses a truncated ResNet-18 and performs direct regression on the relative pose (heading and placement) of the robot to the center of the path, which is then used as a feedback for the controller.
However, these deep-learning based methods require manual annotation of numerous training images, which can be laborious to annotate. Our proposed method utilizes automaticlly generated annotations to train the model, allowing the algorithm to be smoothly deployed in the foreign fields and easily re-calibrated as needed.
Deep reinforcement learning models~\cite{martini2022position,zhou2014vision} have also gained attraction recently. 
However, RL-based approaches require many interactions with the environment and are typically trained only on simulations. RL models also tend to overfit to specific environments.

Finally, most works on in-field autonomous navigation primarily focus on line tracking (walking between rows of plants). 
However, for a robot to navigate fully autonomously in the field, it must also be able to switch rows to next rows of plants after finishing a line.
Some recent works~\cite{ahmadi2022towards,ahmadi2020visual} have proposed methods to perform row switching. 
For example, \cite{ahmadi2020visual} uses the back camera to detect the next row using SIFT detection after finishing a line and then moves backward to the next row. \cite{ahmadi2022towards} applies a similar technique to detect the next row, but employs a robot that can move sideways, thus simplifying the row-switching maneuver.
However, these methods work are only applicable to crop fields where plants laying low on the ground and do not work for vineyards or more complicated fields.  Additionally, the demonstration of row-switching in \cite{ahmadi2020visual} shows that the robot steps/walks over the crop plants during the process, which may damage the crop or not even viable for certain plants, such as those in vineyards.

%\input{dump/related-ref}
%%%%%%%%%%%%%%%%%%%

\section{Method}
%Our proposed navigation algorithm was a state machine consisting two state: row tracking and row switching. The navigation algorithm assumed that the robot was initially pointed down to the row, and started from row tracking. The robot changed to row switching state when the low-precision GPS indicated that the robot was approaching the end of the row. The row switching algorithm drove the robot to align to the next row and then handed over to the row tracking state.

%\begin{figure*}[t]
%  \centering
%  \includegraphics[trim={4cm 6.5cm 3.5cm 5.5cm},clip,scale=0.72]{Figures/Pipeline.png}
%  \caption{Overall design of our vision based navigation approach. The algorithm alternatively switch between row tracking and row switching state to traverse through the field}
%  \label{pipeline}
%\end{figure*}

\begin{figure*}[t]
  \centering
  \includegraphics[trim={1.2cm 8.5cm 5cm 5.8cm},clip,scale=0.65]{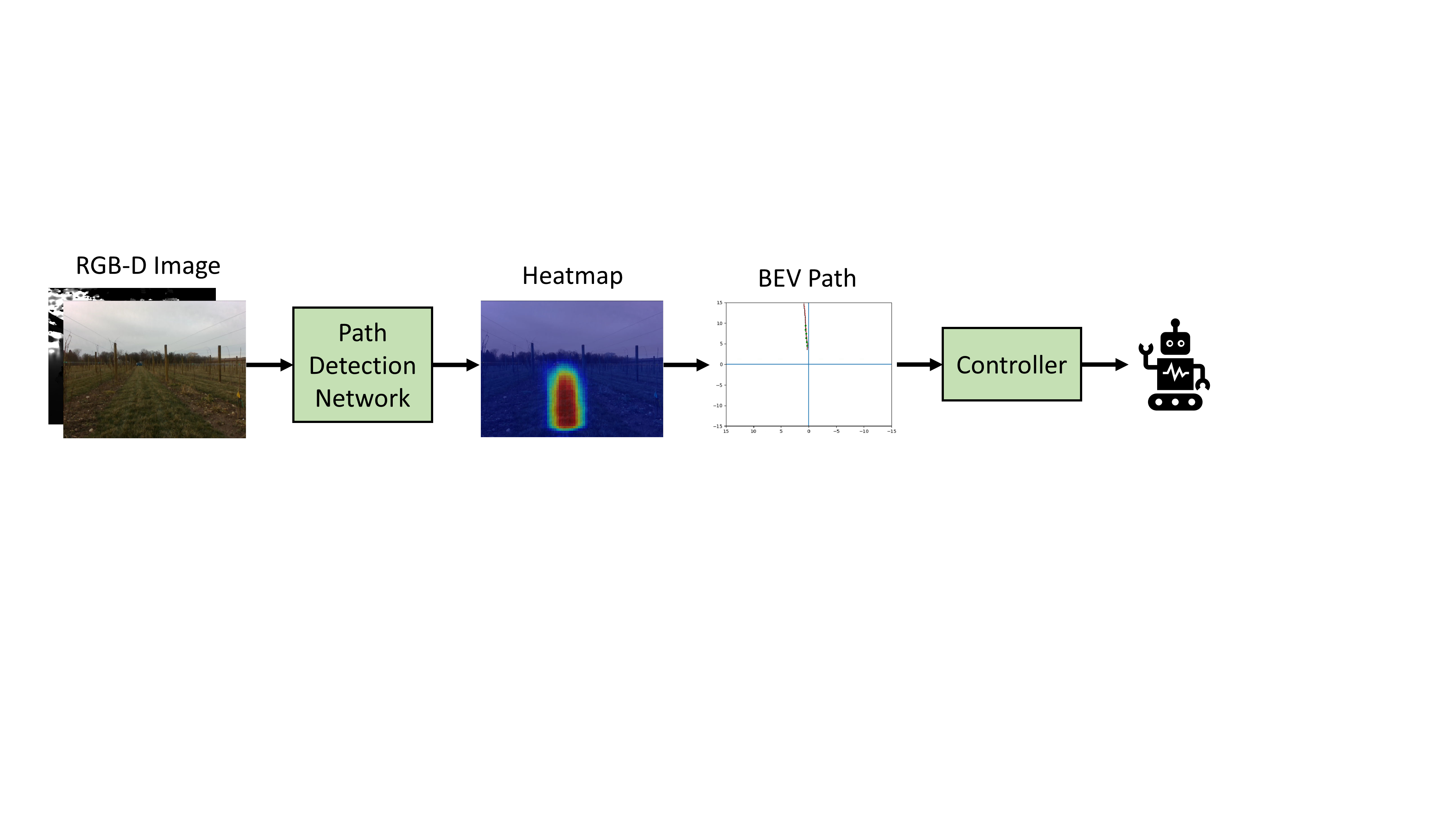}
  \caption{Path detection and BEV controller for row tracking.}
  \label{pipeline1}
\end{figure*}

\begin{figure}[h]
  \centering
  \includegraphics[trim={15cm 7.5cm 1.5cm 0.5cm},clip,scale=0.5]{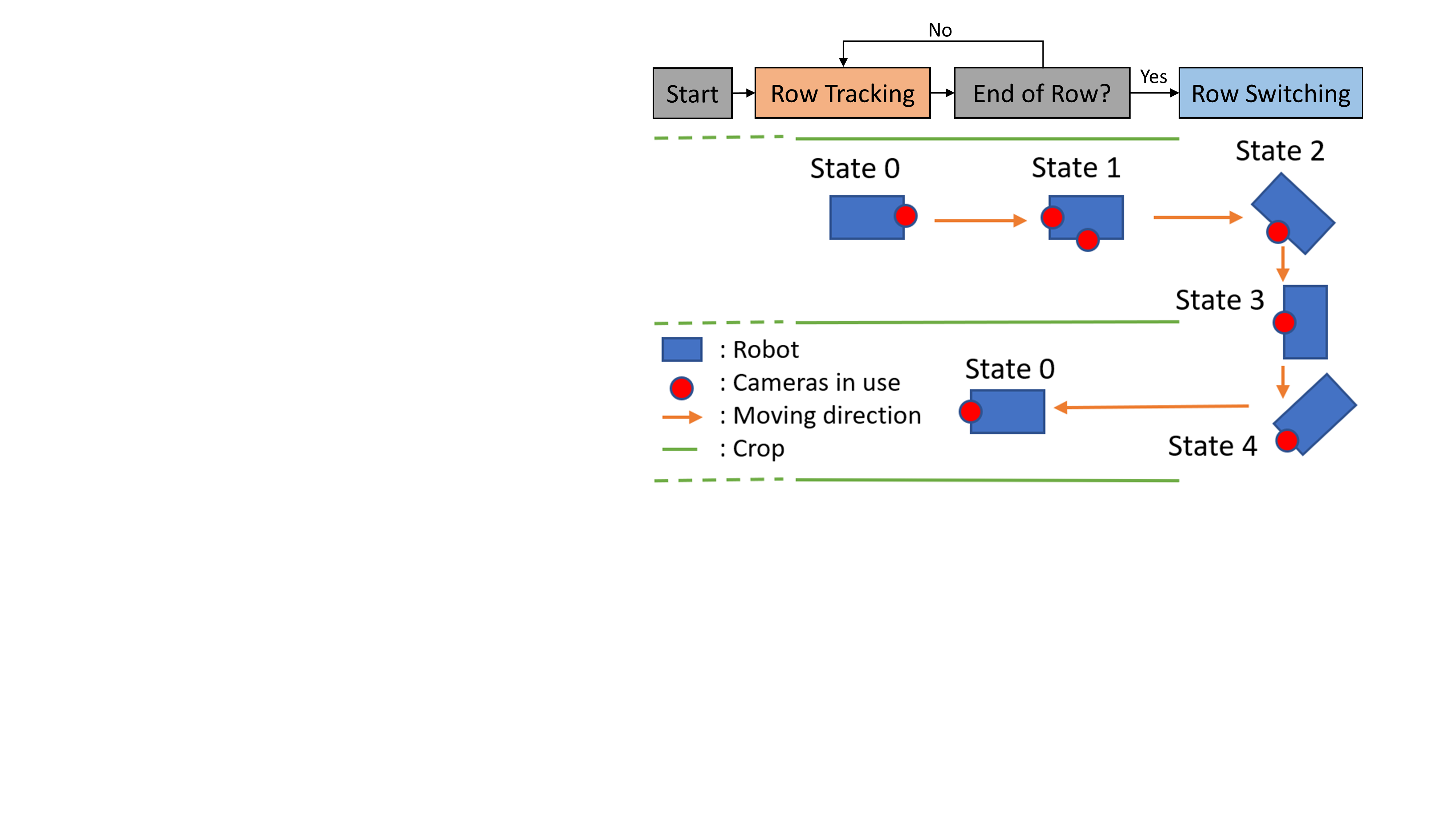}
  \caption{Overall design of our vision based navigation approach. The algorithm alternatively switch between row tracking and row switching state to traverse through the field.}
  \label{pipeline}
\end{figure}

Our autonomous navigation framework comprises two main modules: row tracking and row switching. The framework starts the row tracking module by assuming that a robot will start at the beginning of the first row and head towards the row. As the robot approaches the end of the current row, the framework will use the row switching module to transition the robot into the next row and reverse to the row tracking module. The framework continues this iteration until all rows in a navigation sequence are traversed. 

%\subsection{System design}
%The navigation solution was designed for the differential drive non-holonomic system. Specifically, we integrated our method on a Clearpath Husky based UGV system. The system contained 4 Intel Realsense D435 cameras, one facing foward, one facing backward and two facing each side. The cameras were installed roughly on the centerline of the robot and on a shaft about 1m away from the ground. A SWIFT GPS was added to provide state switching information. All required sensors were connected to an Nvidia Jetson Xavier AGX computer, which execute the algorithm and send the velocity commands to the Husky robot base. The proposed method would work generally in robotic systems that have required equipment and can perform zero-point turns.

\begin{figure}[h]
  \centering
  \includegraphics[width=1\linewidth]{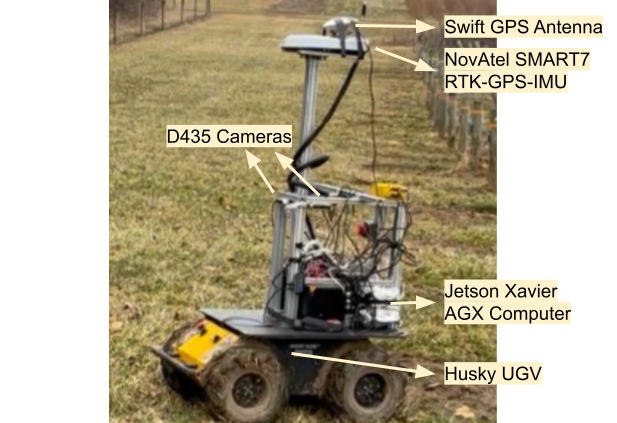}
  \caption{Robotic platform used for data collection, deployment and evaluation. Only two out of four cameras are shown due to viewing angle}
  \label{robot}
\end{figure}

We implemented and evaluated our navigation framework using a custom robot developed for vineyard research and management (Figure \ref{robot}). The robot consisted of a non-holonomic differential drive mobile robot base (Husky, Clearpath Robotics Inc., Canada), four RGB-D cameras (Realsense D435, Intel Corp., USA), and two sets of GNSS systems. The RGB-D cameras were mounted approximately one meter above the ground, with one facing in direction. A low-cost GPS system (Duro, Swift Navigation, USA) was used to provide DGPS signals with up to 5m accuracy to aid in transitioning between row tracking and row switching. A high-end GNSS-IMU system (SMART7, NovAtel Inc., Canada) was used to acquire RTK GPS with centimeter accuracy for training annotation generation and performance evaluation. Both GPS systems worked at 10 Hz. All sensors were connected to an embedded computer (Jetson Xavier AGX, Nvidia Corp., USA) that executed the navigation algorithm and communicated with the robot base for control.

%%%%JM fig%%%%%%%%%%%%%%%%%
%%%%%%%%%%%%%%%%%%%%%%%%%%%

\subsection{Row Tracking}
\subsubsection{Path Detection on Image Frame} \label{sec:Path Estimation on Image Frame}
We proposed a novel learning-based approach to estimate the path of the robot without manual annotation. The path detection in an image was considered as the prediction of a \emph{path traversability heatmap} in which the pixel intensity represented the path preference. The predicted heatmap was used to generate an efficient path for a robot to follow. Pixels with the highest value of individual rows in the heatmap were selected to generate the path in the image frame. The ResNet-18 network \cite{he2016deep} was adapted to estimate the path traversability heatmap of a given RGB-D image. Specifically, the last average pooling layer and all fully convolution layers were replaced with two transposed convolution layers and one single output channel convolution layer (i.e., $M\times N \times1$). This modification enabled the generation of a high-resolution (half size of an input image) heatmap of path preference. Additionally, the first convolution layer of the network was revised to accept four channels (RGB-D) instead of three (RGB), allowing an effective integration of color and depth information for path traversability estimation, especially in complex field conditions. The network was trained using a binary cross-entroy loss between predicted and ground-truth heatmaps.\\

An automatic method was developed to generate ground-truth heatmaps. The robot system would be firstly deployed in a vineyard for collecting RGB-D images from the front camera along with GNSS-IMU information using a waypoint-based navigation with the RTK GPS enabled. The collected GPS paths were projected into the images using the equation \ref{proj_eq}:
\begin{equation}
\boldsymbol{x_c} = \boldsymbol{P} \boldsymbol{[R|t]} \boldsymbol{x_w}
\label{proj_eq}
\end{equation}
where $\boldsymbol{x_w}$ is the GPS path to be projected, $\boldsymbol{[R|t]}$ is the translation between world frame and local camera 3D frame obtained from the GNSS-IMU observation, $\boldsymbol{P}$ is the camera projection matrix, and $\boldsymbol{x_c}$ is the projected path in the image frame

A Gaussian kernel with $\sigma=15$ was applied around the projected paths to generate ground-truth heatmaps for network training. While our approach still required the acquisition of training images, image annotation was fully automatic to make network training and validation much more affordable and scalable. This is a key difference compared with other learning-based methods~\cite{aghi2021deep,sivakumar2021learned} that required laborious annotation of image labels such as canopy/vegetation masks and/or crop row paths or horizon lines.

\begin{figure}[h]
  \centering
  \includegraphics[trim={4.5cm 8.5cm 7cm 5cm},clip,scale=0.42]{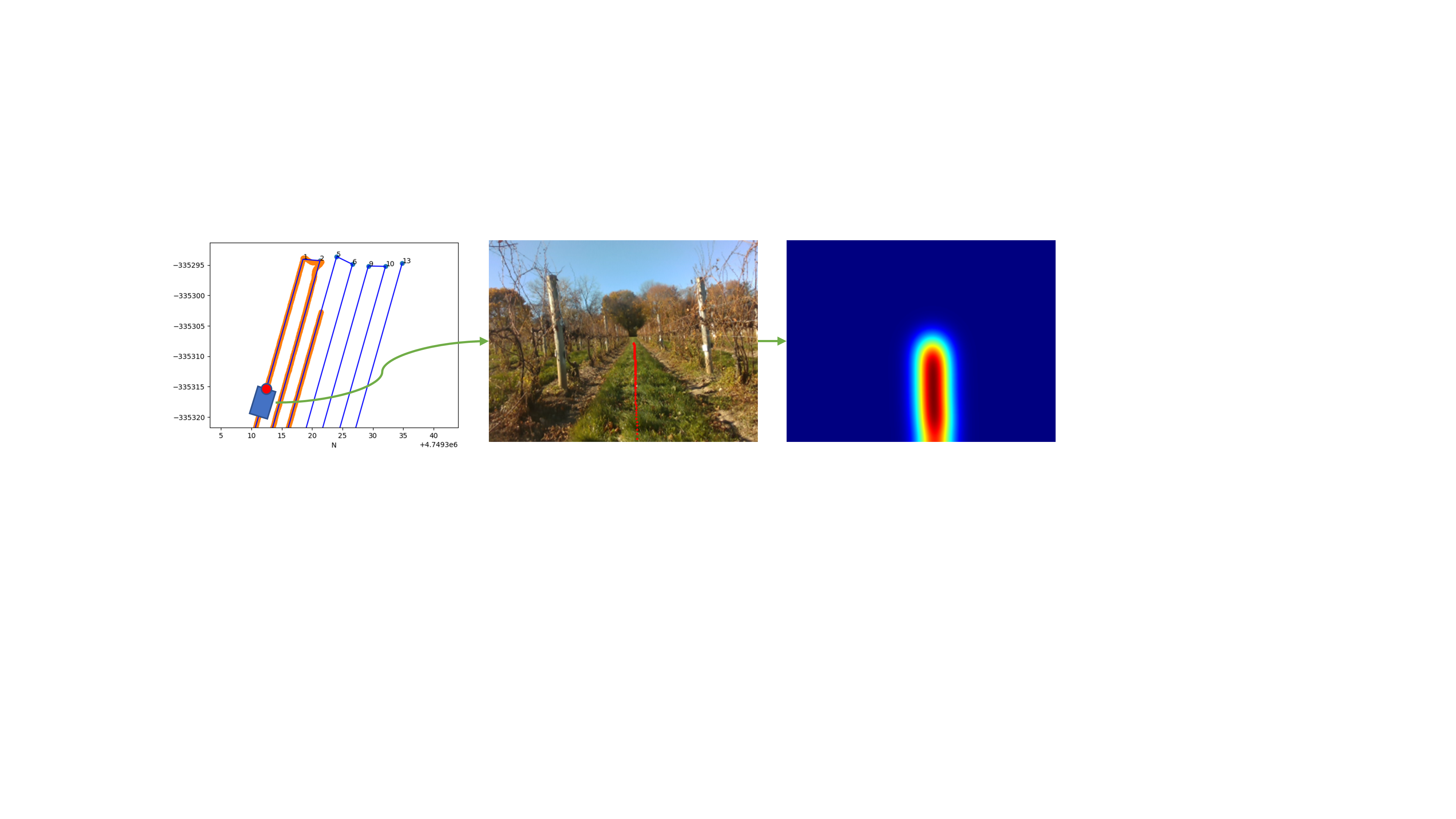}
  \caption{Automatic annotation generation process. The GPS path is projected onto the image frame to generate ground truth label. Image is sampled from the training dataset}
  \label{annotation}
\end{figure}

%Unlike most of the methods described previously, the training annotation of our approach was generated automatically from the RTK-GPS-IMU system with little human effort. To generate the training annotation, the robot first traversed through the pre-determined waypoints in the field using the high precision RTK-GPS-IMU system. The robot position, orientation and camera image were recorded. After the data collection, the GPS paths were projected onto the image using robot positional information and camera projection parameters. In the end, a Gaussian kernel was applied to the projected path on the image frame to generate the ground truthing heatmaps. Although training annotation required RTK-GPS-IMU system, this process only required limited data and only needed to be conducted once when deploying in the foreign field. The robot would navigate independent of the RTK-GPS-IMU system as long as there were no significant changes in the field. 

%We employ a binary cross-entropy loss to train the network.

\subsubsection{BEV Controller}
%During the post-processing stage, the path detected by the network in the image frame was first projected to the Bird Eye view (BEV). The current simplified BEV projection assumed that the camera was always on a fixed height parallel to the ground. This assumption was largely violated due to the rough terrain in the field, but the experimental result indicated that the error was overall neglect-able, and the assumption can be lifted with little modification. After the path was extracted in the BEV, a trajectory follower with a proportional–derivative (PD) controller and feedback Linearization was applied to generate the forwarding and turning velocity commands. Controlling from BEV prevented distorted deviation calculation due to image projection effect.

We designed our controller in the bird-eye's view (BEV) representation rather than in the image space. This was because 1) the BEV representation depicted the robot’s movements with limited effects due to camera pose changes and 2) the BEV representation provided a stable and consistent view of the environment, improving the accuracy and robustness of the controller. Therefore, the extracted path from a given image was transformed to the bird-eye view's representation using homography projection. The resultant path was used as a reference for path following control via a proportional–derivative (PD) controller. Linearization feedback was applied to generate the linear and angular velocity commands for the robot base.

\subsection{Row Switching}
The framework transitioned to the \emph{row switching} module when the robot approached the end of a row. The low-cost GPS was used to roughly estimate the distance between the robot and the pre-surveyed end point of that row. If the distance was less than a threshold, the module transitioning would be triggered to change the robot from state 0 to state 1 in \autoref{pipeline}. In this study, we set an arbitrary threshold of 12 m by considering the GPS accuracy (up to 5 m) and grapevine block length. The use of the low-cost GPS was to relax many difficult requirements that would otherwise hinder the widespread adoption of developed framework and robot. In state 1, the robot started to use the back camera to perform row tracking because the front camera would lose sight of that row. To accurately determine turning point for row switching, one of the side cameras (depending on the turning direction) was used to obtain side view depth distribution. The depth values from the side view increased at the end of a row compared with that in a row.     

Once the turning point was identified, the robot performed the row switching process as follows:
\begin{enumerate}[label=(\alph*)]
    \item After the robot reached the turning point, it executed a 90$^{\circ}$ in-place turn (state 2 in \autoref{pipeline}). 
    Images from the side camera (the same one used for module transitioning) was fed to the path estimation network (described in Section \ref{sec:Path Estimation on Image Frame}) to identify the line of the \emph{recently completed row} (as feedback), from which the relative angle of robot rotation was estimated. The 90$^\circ$ turning concluded once the side camera aligned with the recently completed row, namely, a row path was identified in the center of a side view image.

    \item Now that the robot was considered perpendicular to the grapevine rows and moved straight forward towards the beginning of the next row (state 3 in \autoref{pipeline}). Side camera images were used to identify the line of the \emph{next row} and measure the relative location of the robot to this new row. The robot stopped when it arrived at the \emph{next} row, indicated by the row detected at the \emph{center} of a side view image.

    \item Finally, the robot performed another 90$^{\circ}$ in-place turn to align its orientation with the new row (state 4 of \autoref{pipeline}). This turning process was identical to the first turning (step a), but the front camera (instead of the side camera) was used for feedback because of a better view to the row during the process. After completing this turn, the robot headed towards the new row and switched back to the row tracking module for operations.
\end{enumerate}

\section{Experiment}

\begin{figure*}[t]
  \centering
  \includegraphics[trim={0 0 0 0},clip,scale=1.2]{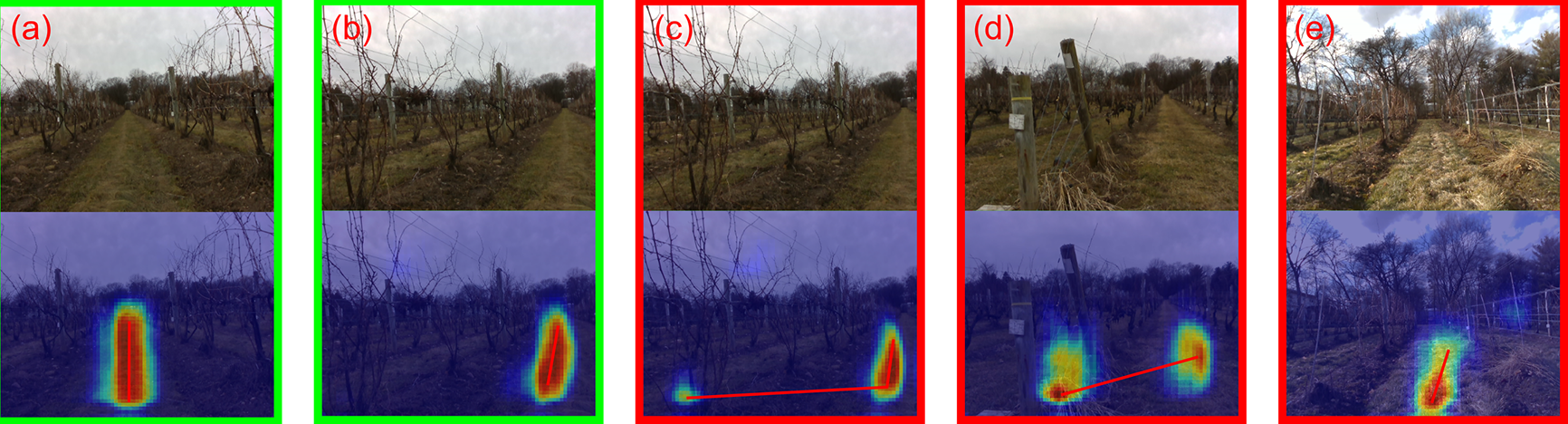}
  \caption{Sample pictures of model inference result. Images of successful detection are labeled in green and images of failed detection are labeled in red. The approximate position and orientation of detected paths are annotated in red.}
  \label{examples}
\end{figure*}

The experiments took place at three different vineyards with \emph{varying conditions} at Cornell AgriTech, Geneva, NY, including Crittenden Pathology vineyard (CRT), VitisGen3 vineyard (VG3), and research north vineyard (RN). CRT and VG3 are on an approximately 10$^{\circ}$ slope and next to each other. While CRT and VG3 share similar terrain and soil conditions, they contain grapevines in very different growth stages (20 years in CRT and newly planted in VG3) and therefore visual appearance. RN is 1.5 km away and has a relatively flat surface and sandy soil. Crop rows in all three vineyards are generally straight lines. The row length is 120m, 90m and 70m for CRT, VG3 and RN, respectively.

%%To train the path detection network, we collected a set of \emph{training data} in CRT field from row 6 up to row 8 (refer to \autoref{trajectory}) in November 2022.
A \emph{training dataset} was established using data from CRT in November 2022 when grapvines still had canopies.
The robot was driven using an RTK-enabled waypoint navigation to collect RGB-D images from the front camera along with RTK-GPS and IMU information. The collected data were used to \emph{automatically} generate a training dataset with ground-truth annotation (as described in Section \ref{sec:Path Estimation on Image Frame}).
Additionally, we introduced deviation from the RTK-guided path when driving the robot to increase the training data variance. Specifically, we manually override the navigation command and change the heading of the robot in the middle of the row for about 10s.
The network was trained using the Adam~\cite{kingma2014adam} optimizer with a batch size of 64, a base learning rate of $1\times 10^{-5}$, and a weight decay of $0.004$. The training process on 4560 RGB-D images (640$\times$480) took approximately 24 hours for 1500 epochs using 2 RTX3090 GPUs. The model weights with the best validation performance were used in all successive tests.

Two tiers of tests were conducted to evaluate the developed navigation framework. The first-tier tests were to evaluate the performance of the path estimation network (Section \ref{sec:Path Detection Evaluation}) as the core component of the framework, and they were perfomed in CRT only. The second-tier tests were to assess the full navigation framework (Section \ref{sec:Autonomous Navigation Evaluation}), and they were conducted in all three vineyards to evaluate the robustness and generalizability of the navigation framework \ref{sec:Generalization Test}. All tests were performed in February 2023 that was three months apart from the training data collection time, furthering the difference between training and testing data to demonstrate the robustness and generalizability of the model and framework.

%The experiments were conducted at the Cornell Pathology Vineyards of Cornell AgriTech in Geneva, NY in the year of 2022 and 2023. The experiments were done in three different fields: crittenden (CRT), vitisgen3 (VG3), and research north (RN). CRT and VG3 are close to each other and are on a roughly 10$^{\circ}$ slope, and RN is about 1.5 km away from them and is relatively flat. The crop rows in all three fields are overall straight with the lengths of around 120m, 90m and 70m for CRT, VG3 and RN respectively.  The training data was collected on November 8th, 2022, in field one. The training data collection started from the east end of row 6 the CRT and ended at the first 10m of row 8. The validation data consisted of the rest of row 8. The robot was driven by the RTK-GPS-IUM based navigation program. Deviations were introduced by manual control override for increasing data variance. During the data collection, the front D435 camera was used, outputting RGB-D image of 640x480 resolution at 6 FPS. The training was conducted on a desktop computer with two Nvidia RTK 3090 GPU and one AMD 3970X CPU. The batch size was set to 64. The Adam optimizer with base learning rate of 1E-5 and weight decay of 0.004 were used. The weight that yielded the lowest loss on the validation dataset was selected. The training process took around 24 hours. After the training, all the evaluation experiments were conducted 3 months later using the same weight.

\subsection{Path Detection Evaluation}\label{sec:Path Detection Evaluation}
Correct identification of a path was the key to the success of the path detection model and successive navigation, especially when a robot oriented away from the row direction. To evaluate this, the robot was manually driven in the middle of a row to perform a $360^{\circ}$ in-place turn and collect inference results from the path detection model. The detected path heading was compared with the robot's heading (from the IMU of the SMART7 system) with respect to the reference path that was generated by pre-surveyed RTK GPS points. 

\setlength{\tabcolsep}{4pt}
\begin{table}[h]
\caption{Path Detection Evaluation Result}
\label{table1}
\begin{center}
\begin{tabular}{cccccc}
\toprule
Robot Heading ($^{\circ}$) & \text{[}-25, -15\text{)} &  \text{[}-15, -5\text{)} & \text{[}-5, 5\text{)} & \text{[}5, 15\text{)} & \text{[}15, 25\text{]} \\

\midrule
Avg Deviation ($^{\circ}$) &2.36&2.22&1.10&1.57&8.05\\
\bottomrule
\end{tabular}
\end{center}
\end{table}

The model could provide satisfactory path detection for a robot heading up to $\pm25^{\circ}$ away from the row direction (Table \ref{table1}). When the robot heading was beyond this range, the traversal path was not fully visible in images, yielding unusable results. For most of the valid heading angles, the average heading deviation was below $2.5^{\circ}$, which was comparable to a previous study ($1.99^{\circ}$ average heading error) on vision navigation in a corn field~\cite{sivakumar2021learned}. The average deviation generally increased as the angle between the robot heading and the crop row increased. At larger angles, the collected images started to include information from adjacent rows, resulting in false detection. This issue may be partially addressed during a growing season when grapevine canopies would block the view of adjacent rows.

These patterns were observed from representative images (Figure \ref{examples}). 
When the robot headed straight down to the row (Figure \ref{examples} a) or within the heading deviation range ($\sim20^{\circ}$ away from the row in Figure \ref{examples} b), the model could accurately generate a high quality heatmap and identify the correct moving path. However, when the robot headed out of the deviation range (i.e., greater than $25^{\circ}$), one side of the row became invisible in the image, and the model started to be influenced by the features from the next row (Figure \ref{examples} c). In addition to the heading deviation effect, tall weeds and vacant space due to dead grapevines introduced challenges for the model to correctly detect the traversable path (Figure \ref{examples} d and Figure \ref{examples} e). While these special cases were occasional in vineyards and generally did not hamper the continuation of navigation, some large navigation errors were observed due to suboptimal path detection results. These could be mostly improved by fine-tuning the model with the automatic annotation without significant additional effort.
%%%%%%%%%%%%%%%%%%%%%%%%%%%%%

\subsection{Autonomous Navigation Evaluation}\label{sec:Autonomous Navigation Evaluation}
Field trials were conducted on February 16, 2023 in all three vineyards to quantitatively analyze the navigation framework performance. In each vineyard, four replication trials were conducted using the developed vision-based navigation framework to simulate the periodical and repetitive fieldwork needs. For the vision-based navigation, all cameras were set to the resolution of 640 $\times$ 480 at 15 FPS. During the trial, the RTK-GPS locations were separately logged for trajectory analysis purposes. One reference trial with RTK-enabled waypoint navigation was conducted for comparison purposes.

In CRT, the trial started from the end (east side) of row 7 and stopped at the first 10m (east side) of row 11, consisting of 4 full row tracking maneuvers and 4 row switching maneuvers (2 on each side of the vineyard). Although row 7 and 8 were included in the training and validation dataset that was collected 3 months ago, the robot traversed through these two rows from the opposite directions with different visual features. 

Positional and heading deviations between robot moving trajectories and predefined crop row paths were measured and used as metrics for performance evaluation. The positional deviation was defined as the distance from the current robot position to the predefined crop row path, and the heading deviation was defined as the robot heading with respect to the crop path direction. In these analyses, the robot heading was calculated using GPS records. To avoid potential noises in the calculation, the collected GPS records were downsampled to 1Hz. To quantify the algorithm performance at row tracking and row switching stages individually, the data near the end of the rows were separately analyzed from the data in the middle of the rows. The end of the row region was defined by 12m radians circles from the crop row endpoints.

\begin{table*}[t]
\caption{Trajectory analysis for navigation in CRT}
\label{table2}
\begin{center}
\begin{tabular}{cccccccc}
\toprule
{} &  {} &  \multicolumn{3}{c}{Positional deviation (mean $\pm$ std, max)(m)} & \multicolumn{3}{c}{Heading deviation (mean $\pm$ std, max)($\circ$)}\\\cmidrule{3-8}
{} & \#Intervention & Row tracking   & Existing    & Entering & Row tracking   & Existing    & Entering\\

%\multirow[b]{2}{*}{xxxxxxxxxxx aaa} & \multicolumn{2}{c}{SPF (s)} & \multicolumn{2}{c}{mIoU (\%)}\\\cmidrule{2-5}
%   & PyTorch   & TensorRT    & PyTorch & TensorRT\\

\midrule
Trial1 &0&0.19 $\pm$ 0.13, -0.52&0.22 $\pm$ 0.11, -0.45&0.27 $\pm$ 0.16, -0.50&1.94 $\pm$ 1.69, 14.53&2.70 $\pm$ 2.24, 12.78&2.14 $\pm$ 1.71, 7.25 \\
Trial2 &1&0.18 $\pm$ 0.11, -0.50&0.19 $\pm$ 0.11, -0.40&0.18 $\pm$ 0.16, -0.47&1.68 $\pm$ 1.33, 9.27&2.88 $\pm$ 2.37, 11.80&1.96 $\pm$ 1.83, 8.11\\
Trial3 &0&0.18 $\pm$ 0.11, -0.50&0.22 $\pm$ 0.21, -0.80&0.28 $\pm$ 0.17, 0.67&1.87 $\pm$ 1.53, 13.47&2.89 $\pm$ 2.21, 9.71&2.49 $\pm$ 2.12, 9.93\\
Trial4 &1&0.17 $\pm$ 0.11, -0.47&0.27 $\pm$ 0.25, -1.02&0.15 $\pm$ 0.10, -0.33&1.79 $\pm$ 1.48, 9.80&2.99 $\pm$ 2.49, 10.29&2.00 $\pm$ 1.83, 8.33\\
RTK-GPS &0&0.01 $\pm$ 0.01, -0.05&-&-&1.18 $\pm$ 0.89, 5.03&-&-\\
\bottomrule
\end{tabular}
\end{center}
\end{table*}

\begin{figure*}[t]
  \centering
  \includegraphics[trim={3.5cm 5cm 4cm 4.8cm},clip,scale=0.75]{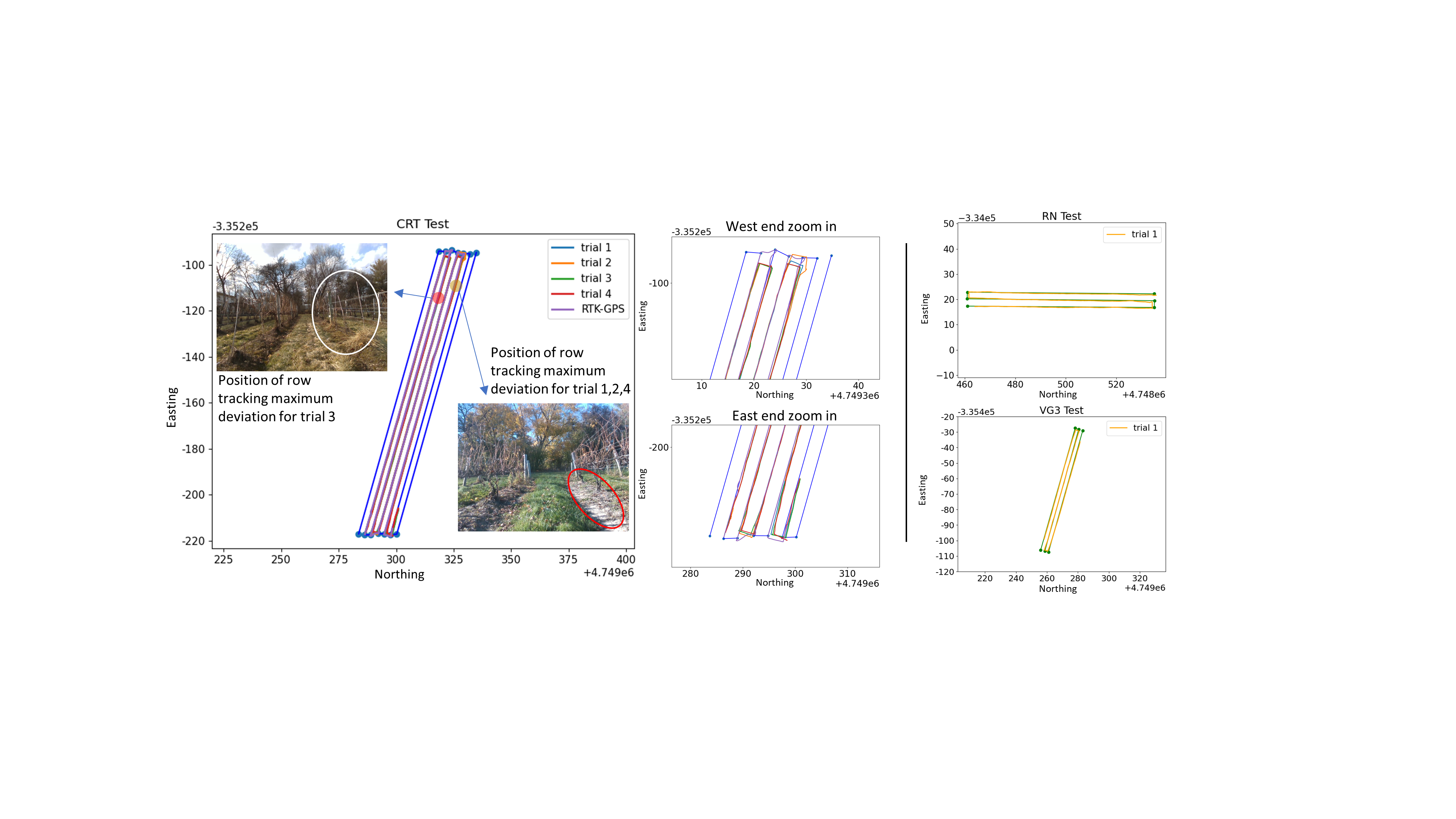}
  \caption{Autonomous navigation evaluation and generalization evaluation trajectories. Left: Five trials in CRT with positions of maximum deviations of vision based row tracking labeled. Right: Generalization evaluation trajectories in RN and VG3. Easting coordinates are flipped along the Northing axis.}
  \label{trajectory}
\end{figure*}

Among the four replication trials, two trials were successfully completed without any human intervention, whereas another two trials required human intervention once for correct row switching to complete the trials (\autoref{table3} and left panel in \autoref{trajectory}). This demonstrated that the developed navigation framework generally provided a satisfactory performance, especially for row tracking (no human intervention along a total of 1920 m testing distance). However, row switching presented challenges for the current framework occasionally. In trial 2, during the transitioning from row 9 to row 10 (west end), the noisy depth information from the side camera failed to initiate the row switching module to state 1, and therefore a human intervention was needed to correctly turn the robot for successive navigation. In trial 4, during the transitioning from row 10 to row 11 (east end), the robot deviated too far away from the optimal path while navigating using the back camera, and as a result the robot mis-detected the next row as the current row in state 3. A human intervention was required to direct the robot back to the correct row.

The positional and heading deviation distributions were similar cross the 4 trials, indicating stable performance for repetitive runs on the same path (\autoref{table3}). While the position deviations for the vision based navigation were higher than those of the RTK-enabled waypoint navigation, the heading deviations were comparable. This was likely due to the fact that angular errors were more visible in image frames especially at further distances to the camera, while the positional offset between the image center-line and the crop row center-line tended to vanish as the distance increased due to image projection. The deviations during row switching (especially when the robot entered or exited a row) were also higher than those during row tracking deviations because of uncertainties caused by the path detection using the back camera. When tracking using back camera, the BEV controller needed to extend the detected path to the front of the robot. This extension increased the precision requirement for the detection and decreased the stability of the controller. This issue could be solved by improving the detection accuracy and the controller design.

While large positional deviations occurred very occasionally during the field trials, the maximal positional deviations were observed for common challenges (Left panel in Figure \ref{trajectory}). For trial 1, 2, and 4, the maximum deviation happened at roughly the same location where a ditch was created by previous tractor operations (red circle in the left panel in Figure \ref{trajectory}), introducing terrain differences for robot maneuver. For trial 3, the maximum deviation happened around a vacant space where several adjacent grapevines were dead and removed (white circle in the left panel in Figure \ref{trajectory}). This vacant space could influence the path detection model to generate a suboptimal path with a large positional deviation.     

Overall, the trial results demonstrated that our vision-based navigation framework was able to continuously guide the robot traversing through a vineyard with minimum human intervention, even during the winter time that presented additional terrain challenges. With extensive grapevine canopies (therefore better visual and depth information of a crop row) during the growing season and improved row switching, one could expect the framework to continuously perform well as an all-season navigation solution in the future. 

%%%%%%%%%%%%%%%%%%%%%%%%%%%%%
\subsection{Generalization Test}\label{sec:Generalization Test}
To evaluate the robustness and generalizability of the navigation framework in unseen vineyards, additional field trials were conducted on February 17th, 2023 in VG3 and RN. For each vineyard, the trial included 3 row tracking maneuvers and 2 row switching maneuvers (1 on each end of the field). Only one trial was performed with vision-based navigation. The data logging and evaluation processes were identical to those in CRT field trials.

\begin{table*}[t]
\caption{Trajectory analysis for navigation in RN and VG3.}
\label{table3}
\begin{center}
\begin{tabular}{cccccccc}
\toprule
{} &  {} &  \multicolumn{3}{c}{Positional deviation (mean $\pm$ std, max)(m)} & \multicolumn{3}{c}{Heading deviation (mean $\pm$ std, max)($\circ$)}\\\cmidrule{3-8}
{Field} & \#Intervention & Row tracking   & Existing    & Entering & Row tracking   & Existing    & Entering\\

%\multirow[b]{2}{*}{xxxxxxxxxxx aaa} & \multicolumn{2}{c}{SPF (s)} & \multicolumn{2}{c}{mIoU (\%)}\\\cmidrule{2-5}
%   & PyTorch   & TensorRT    & PyTorch & TensorRT\\

\midrule
RN &0&0.19 $\pm$ 0.09, -0.40&0.25 $\pm$ 0.19, -0.75&0.29 $\pm$ 0.06, 0.38&1.78 $\pm$ 1.40, 7.22&3.71 $\pm$ 3.50, 16.71&1.65 $\pm$ 1.46, 7.40\\
VG3 &0&0.19 $\pm$ 0.10, -0.40&0.31 $\pm$ 0.15, -0.66&0.32 $\pm$ 0.16, 0.66&1.81 $\pm$ 1.44, 8.67&3.04 $\pm$ 1.85, 6.58&3.19 $\pm$ 2.50, 9.90 \\
\bottomrule
\end{tabular}
\end{center}
\end{table*}

%Trial 1,"0.19,0.13,-0.52","0.22,0.11,-0.45","0.27,0.16,-0.50","1.94,1.69,14.53","2.70,2.24,12.78","2.14,1.71,7.25"
%Trial 2&0.18,0.11,-0.50&0.19,0.11,-0.40&0.18,0.16,-0.47&1.68,1.33,9.27&2.88,2.37,11.80&1.96,1.83,8.11
%Trial3&0.18,0.11,-0.50&0.22,0.21,-0.80&0.28,0.17,0.67&1.87,1.53,13.47&2.89,2.21,9.71&2.49,2.12,9.93
%Trial 4&0.17,0.11,-0.47&0.27,0.25,-1.02&0.15,0.10,-0.33&1.79,1.48,9.80&2.99,2.49,10.29&2.00,1.83,8.33
%RTK-GPS,"0.01,0.01,-0.05",-,-,"1.18,0.89,5.03",-,-

Both trials successfully completed without any human intervention (Table \ref{table3} and Figure \ref{trajectory}) and achieved comparable positional and heading deviations to those in CRT, indicating a consistent satisfactory performance in unseen vineyards without model fine-tuning and parameter re-tuning. It should be noted that RN is oriented in the North-South direction, which is different from CRT and VG3 along the East-West direction. Operating in RN required the camera to directly facing the Sun. The experiment results indicated that no significant interference was caused during row tracking. However, the robot spent longer time to turn into the third row facing the Sun. This was likely caused by the delayed auto-exposure adjustment and could be potentially solved by optimizing the imaging system.

\section{Conclusion}
We developed a novel learning, vision-based framework for autonomous vineyard navigation that lift the requirement of costly human annotation. A training dataset with annotation was automatically generated by a single-time autonomous field acquisition using a RGB-D camera and a RTK GPS. The generated training dataset was used to train a custom network for path detection for row tracking. The same model was also used to perform row switching, allowing the robot to traverse through the field of multiple rows autonomously. The experimental results demonstrated 1) the automatic annotation significantly reduced the cost and lowered the barriers of using learning-based model for navigtaion path detection and 2) the trained network and developed framework achieved satisfactory accuracy, robustness, and generalizability for autonomous navigation even in totally unseen vineyards.

In future studies, we will focus on 1) improving the stability of the row switching module, espcially under challenging cases and 2) adding the centerline tracking feature to the navigation framework for collecting uniform data for plant phenotyping and vineyard management purposes. We will also conduct additional field tests in different cropping systems (e.g., row crops) to explore the full potential of the developed framework.
\section{Acknowledgement}
This work was jointly supported by the USDA National Institute of Food and Agriculture Hatch projects (Accession no. 1024574 and 1025032) and Cornell Institute for Digital Agriculture Seed Fund.

{\small
\bibliographystyle{unsrt}
\bibliography{sections/ref}
}

\end{document}